%% file: main.tex
\documentclass[runningheads]{llncs}
\usepackage[T1]{fontenc}

\usepackage{graphicx}

\usepackage{hyperref}

\usepackage{amsfonts}
\usepackage{amsmath}
\usepackage{booktabs}
\usepackage[misc]{ifsym}
\usepackage{microtype}
\usepackage{multirow}
\usepackage{orcidlink}
\usepackage{threeparttable}
\usepackage{xspace}
\newcommand{\datasetname}{EventEA\xspace}
\usepackage{longtable}

\begin{document}
\title{EventEA: Benchmarking Entity Alignment for Event-centric Knowledge Graphs}
\titlerunning{EventEA: Benchmarking Entity Alignment}
%
\author{
    Xiaobin Tian\and
	Zequn Sun\orcidlink{0000-0003-4177-9199} \and
	Guangyao Li\and
	Wei Hu\orcidlink{0000-0003-3635-6335}\textsuperscript{(\Letter)}}
\authorrunning{X. Tian et al.}
%
\institute{
	State Key Laboratory for Novel Software Technology, Nanjing University, China \\
	National Institute of Healthcare Data Science, Nanjing University, China \\
	\email{\{xbtian,zqsun,gyli\}.nju@gmail.com, whu@nju.edu.cn}
}
\maketitle              
\begin{abstract}
Entity alignment is to find identical entities in different knowledge graphs (KGs) that refer to the same real-world object.
Embedding-based entity alignment techniques have been drawing a lot of attention recently because they can help solve the issue of symbolic heterogeneity in different KGs.
However, in this paper, we show that the progress made in the past was due to biased and unchallenging evaluation.
We highlight two major flaws in existing datasets that favor embedding-based entity alignment techniques,
i.e., the isomorphic graph structures in relation triples and the weak heterogeneity in attribute triples.
Towards a critical evaluation of embedding-based entity alignment methods,
we construct a new dataset with heterogeneous relations and attributes based on event-centric KGs.
We conduct extensive experiments to evaluate existing popular methods, and find that they fail to achieve promising performance.
As a new approach to this difficult problem, we propose a time-aware literal encoder for entity alignment.
The dataset and source code are publicly available to foster future research.
Our work calls for more effective and practical embedding-based solutions to entity alignment.

\medskip
\textbf{Resource Type}: Dataset and benchmarking evaluation

\smallskip
\textbf{License:} GPL-3.0 License

\smallskip
\textbf{DOI:} \url{https://doi.org/10.6084/m9.figshare.19720222.v1}

\smallskip
\textbf{GitHub Repository}: \url{https://github.com/nju-websoft/EventEA}

\end{abstract}

\input{sec1_intro}
\input{sec2_preliminaries}
\input{sec3_dataset}
\input{sec4_method}

\input{sec5_experiment}
\input{sec6_conclusion}
\bibliographystyle{splncs04}
\bibliography{reference}

\end{document}


\author{
    Xiaobin Tian\and
	Zequn Sun\orcidlink{0000-0003-4177-9199} \and
	Guangyao Li\and
	Wei Hu\orcidlink{0000-0003-3635-6335}\textsuperscript{(\Letter)}}
	
\institute{
	State Key Laboratory for Novel Software Technology, Nanjing University, China \\
	National Institute of Healthcare Data Science, Nanjing University, China \\
	\email{\{xbtian,zqsun,gyli\}.nju@gmail.com, whu@nju.edu.cn}
}
\title{EventEA: Benchmarking Entity Alignment for Event-centric Knowledge Graphs}

\titlerunning{EventEA: Benchmarking Entity Alignment}
%
%
%
%
\maketitle              

\section{Hyperparameter Configuration}
\label{ap: parameter setting}

The detailed ranges of hyperparameters for \modelname are summarized in Table~\ref{tab:range}.
We determine the hyperparameter by grid search based on the Hits@1 performance, and we also adopt early stopping strategy.
For other methods implemented by OpenEA, we refer to the default configuration and adjust some common hyperparameters, such as batch size and learning rate. 
\begin{table}[h]\setlength\tabcolsep{8pt}
    \centering
    \caption{Hyperparameter ranges and values in training of \modelname. Here, $\beta$ is in Eq. (4) and $\lambda$ is in Eq. (6) of the paper.
    The selected values are marked in boldface.}
    \label{tab:range}
    \begin{tabular}{lc}
    \toprule
        Hyperparameters          & Values \\
        \midrule 
        Embedding dimension         & 768 \\
        Batch size       & $\{\textbf{256}, 512, 1024\}$ \\
        Optimizer        & Adam \\
        Learning rate     & $\{0.00001, \textbf{0.0001}, 0.001\}$ \\
        $\lambda$                & $\{0.5, 1.5, \textbf{3}, 3.5, 4.5, 5\}$ \\
        $\beta$        & $\{0.01, \textbf{0.02}, 0.05, 0.1\}$ \\
    \bottomrule
    \end{tabular}
\end{table}

%% file: sec1_intro.tex
\section{Introduction}\label{sect:intro}
Entity alignment, seeking to find identical entities referring to the same real-world in different knowledge graphs (KGs), has a long research history.
Early work on entity alignment focused primarily on the literal similarity of entity attributes or logical reasoning based on relation semantics \cite{PARIS}. 
Recent research attention is turned to embedding techniques that encode similar entities with similar vector representations \cite{MTransE,JAPE,BootEA}.
The resulted joint KG embeddings not only capture entity alignment but also allow knowledge transfer between multiple sources to benefit downstream tasks \cite{MultilingualKBC}.
Entity alignment is a non-trivial task because the relation and attribute triples of an entity are usually incomplete or even unavailable.
Although embedding-based methods have made significant progress in recent years~\cite{openea}, we need to carefully point out that existing datasets for evaluation have two biases that favor embedding-based methods.

\textbf{Bias 1:} The two KGs in the entity alignment datasets have many isomorphic graph structures.
We have noticed that more and more studies focus on using the relational graph structures to learn entity embeddings, and many of them adopt graph neural networks (GNNs) as a backbone.
However, these methods are all built with a foundational assumption that similar entities have similar subgraph structures \cite{AliNet}.
These entity alignment methods are referred to as structure-based.
We found that existing datasets such as DBP15K \cite{JAPE} and OpenEA \cite{openea} all have isomorphic graph structures in the source and target KGs to be aligned.
For example, the KGs in DBP15K are extracted from the multi-lingual infobox-based triples in DBpedia \cite{DBpedia} and thus have similar schemata.
As a result, the structure-based methods may achieve high performance on these datasets because of the expressive power of GNNs in recognizing isomorphic subgraph structures \cite{GNN_power}.
However, we argue that not all KGs in the real world have similar graph structures due to knowledge incompleteness or schema heterogeneity.
It is still not clear how well these structure-based methods work when they encounter KGs with very few isomorphic subgraph structures.

\textbf{Bias 2:} The attributes and literals in existing entity alignment datasets are less heterogeneous.
In existing datasets such as OpenEA, attribute triples serve as the side information to assist entity alignment.
We also discovered that many embedding-based entity alignment methods \cite{MuiltiKE,IMUSE,AttrE} attempt to incorporate entity attributes, particularly names, into the embedding methods to improve performance.
A widely-used way of incorporating attributes is to encode them as vectors based on pre-trained language models (e.g., BERT~\cite{BERT}) or word embeddings (e.g., FastText~\cite{fasttext}). 
The vectors are then used by the following GNNs to learn entity embeddings as the initial representations \cite{RDGCN,GMNN}.
These methods are known as attribute-enhanced, and they achieve cutting-edge performance on existing datasets.
However, recent studies \cite{AttrGNN} show that many attribute-enhanced methods are overestimated because the attribute values (e.g., entity names) in existing datasets are less heterogeneous.
For example, in DBpedia~\cite{DBpedia} and YAGO \cite{YAGO},
identical entities typically have the same name, which actually leaks the entity alignment information to the embedding method.
Even in cross-lingual entity alignment datasets like DBP15K \cite{JAPE}, the multilingual issue can also be easily resolved by machine translation or multilingual word embeddings.
The root cause of this problem is that the literal heterogeneity in existing datasets is insufficient because they are extracted from the same source, such as the multilingual Wikipedia.

\begin{figure}[!t]
	\centering
	\includegraphics[width=0.999\linewidth]{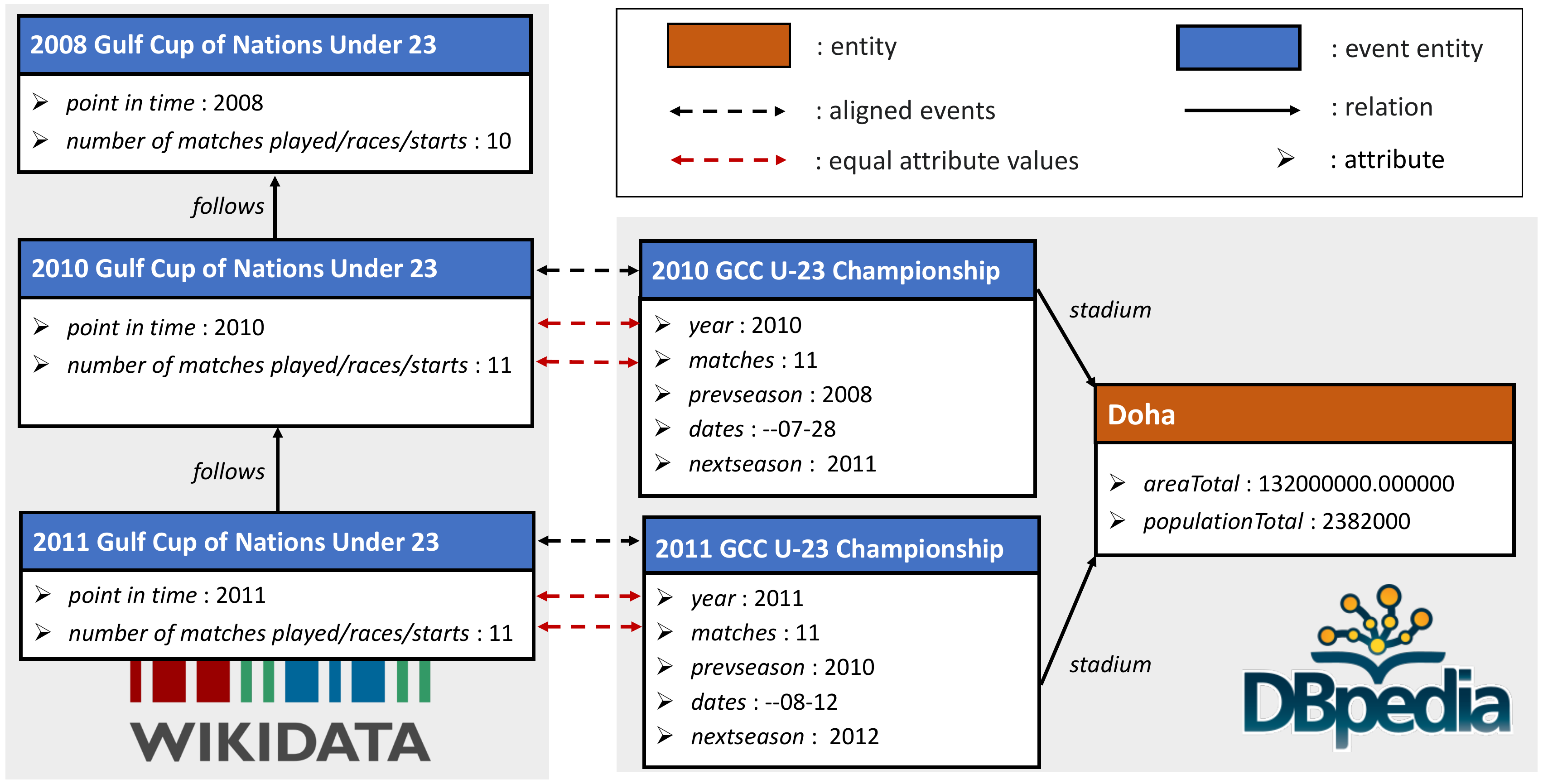}
	\caption{An example of event alignment between Wikidata and DBpedia.}
	\label{fig:event}
\end{figure}

In this paper, we propose a more challenging dataset for embedding-based entity alignment, with the goal of resolving the aforementioned biases.
Unlike previous datasets extracted from open-domain KGs, we choose event KGs for their specific properties that are not suitable for embedding-based entity alignment.
An event KG is a special KG that its entities typically describe events.
As shown in Fig.~\ref{fig:event}, events have multi-dimensional information such as time, location, participants, etc, and event entity alignment is more challenging for the following reasons.
First, the relations between events are sparse, making the graph structures of different event KGs less isomorphic.
According to our observations, most of the relations between events are sequential. 
As shown in Fig.~\ref{fig:event}, in Wikidata, the event entity \textit{2010 Gulf Cup of Nations Under 23} only has a relation with its serial events. 
And in DBpedia, such a relation between the corresponding events is missing due to its incompleteness. 
Instead, they are connected through the location entity \textit{Doha}.
Such sparse KGs with less isomorphic graph structures are a good choice for assessing the robustness of structure-based entity alignment methods.
Second, the names of events are more complicated, and the names of identical events can be greatly different. 
For example, in Fig.~\ref{fig:event}, the entities
\textit{2010 GCC U-23 Championship} from DBpedia and \textit{2010 Gulf Cup of Nations Under 23} from Wikidata refer to the same event but their names have obvious differences. 
Besides, events are usually rich in attributes.
For example, we found more than 30 attributes of \textit{2010 GCC U-23 Championship} in DBpedia.
With more attributes and more heterogeneous values, event entity alignment is better suited than existing datasets for evaluating whether attribute-enhanced methods truly capture the semantic similarities of entity attributes and names.

In summary, our main contributions in this paper are threefold:

\begin{itemize}
\item For evaluating embedding-based methods, we constructed an event-centric entity alignment dataset, namely \datasetname,
by harvesting from EventKG \cite{EventKG}, DBpedia \cite{DBpedia} and Wikidata \cite{Wikidata}.
Compared with existing entity alignment datasets, \datasetname has fewer isomorphic structures but more heterogeneous attribute values.
\datasetname is a good choice for evaluating the robustness of embedding-based methods in dealing with difficult entity alignment settings.

\smallskip
\item We conducted extensive experiments to evaluate existing embedding-based entity alignment methods on \datasetname. 
The results show that structure-based methods fail to achieve promising performance with fewer isomorphic structures, 
while attribute-enhanced methods suffer greatly from the literal heterogeneity (especially the heterogeneous names).
Our experimental findings suggest that we should revisit the progress made by embedding techniques in entity alignment, and more robust methods are needed.

\smallskip
\item To take the lead in problem solving, we propose a time-aware literal encoder in this paper to resolve literal heterogeneity in attribute values.
It aims to emphasize the significance of timestamps in event attribute values such as entity name in order to assist embeddings in capturing the semantic similarities of events.
It outperforms existing embedding-based entity alignment methods on \datasetname.
Our source code and dataset are publicly available.
\end{itemize}

%% file: sec2_preliminaries.tex
\section{Background Knowledge}
We first introduce the preliminaries on entity alignment between event-centric KGs. 
Then we review related work on entity alignment methods and datasets.

\subsection{Preliminaries}
We begin with the definition of a KG.

\begin{definition}[Knowledge graph] 
We define a KG as a six-tuple, i.e., $\mathcal{G}=\{\mathcal{E}, \mathcal{R}, \mathcal{A}, \mathcal{V}, \mathcal{T}_{\text{rel}}, \mathcal{T}_{\text{att}}\}$,
where $\mathcal{E}$, $\mathcal{R}$, $\mathcal{A}$ and $\mathcal{V}$ denote the sets of entities, relations, attributes and literal values, respectively.
$\mathcal{T}_{\text{rel}} \subseteq \mathcal{E} \times \mathcal{R} \times \mathcal{E}$ denotes the set of relation triples.
$\mathcal{T}_{\text{att}} \subseteq \mathcal{E} \times \mathcal{A} \times \mathcal{V}$ denotes the set of attribute triples.
\end{definition}

In this paper, we focus on event-centric KGs \cite{EventKG}.
An event entity possesses certain event-specific features, like time, location, and participants.
A KG built based on event entities is known as an event-centric KG.
In an event-centric KG, event entities account for a large proportion, and they are attached with other entities, such as location entities \cite{EventKG_Survey}.
As we have mentioned in Sect.~\ref{sect:intro}, event-centric KGs present more challenges for entity alignment.
Please keep in mind that, unlike some other studies \cite{MTransE,JAPE,RDGCN}, we do not require each entity to have at least one relation or attribute triple.


\begin{definition}[Entity alignment] 
Given a source KG $
\mathcal{K}_1$ and a target KG $\mathcal{K}_2$,
entity alignment is the task of finding identical entities among their entities $\mathcal{E}_1$ and $\mathcal{E}_2$, 
i.e., $\mathcal{M}=\{(e_1, e_2)\in \mathcal{E}_1 \times \mathcal{E}_2\,|\, e_1 \equiv e_2 \}$, where $\equiv$ denotes the \text{owl:sameAs} relation.
In the supervised setting, a small set of seed entity alignment is provided as training data,
and the task is to find the remaining alignment for other entities.
\end{definition}

This is a general definition and it also applies to our entity alignment task between event-centric KGs.
Compared with existing embedding-based entity alignment settings, 
our task is complicated in the data characteristics of event-centric KGs, which include sparse relational structures and heterogeneous attribute values.
Hence, our task is more realistic and also more challenging.

\subsection{Related Work}

\subsubsection{Embedding-based entity alignment methods.}
The basic idea behind embedding-based entity alignment is to map the source and target KGs into an embedding space and use the embeddings to measure entity similarities for alignment retrieval.
It has developed very rapidly in recent years.
Please refer to recent surveys \cite{openea,EA_survey_AIOpen,TKDE_EA_survey} to gain an overview.
Existing studies generally fall into two groups, i.e., structure-based and attribute-enhanced methods.
The first group of methods \cite{MTransE,RSN4EA,AliNet,GCN-Align,IPTransE} focuses on geometric representation learning techniques, such as translational embeddings \cite{TransE} and GNNs \cite{GCN,R-GCN}, to capture the relational structures of KGs to identify similar entities.
These methods require that each entity must have at least one relation triples for embedding learning.
The second group of methods \cite{KDCoE,IMUSE,AttrGNN,JAPE,AttrE,RDGCN,GMNN} makes use of attribute triples as side information to enhance embedding learning.
However, due to the name bias issue in existing datasets \cite{AttrGNN},
many of these methods use entity names as an opportunistic feature.
They simply encode entity names with BERT \cite{BERT} or FastText \cite{fasttext} and the name embeddings can achieve very high accuracy on some datasets such as DWY100K \cite{BootEA}.
However, these methods are less robust, and this phenomenon motivates our work.

\subsubsection{Embedding-based entity alignment datasets.}
There are four widely-used datasets for evaluating embedding-based entity alignment methods: WK3L \cite{MTransE}, DBP15K \cite{JAPE}, DWY100K \cite{BootEA} and OpenEA \cite{openea}.
WK3L \cite{MTransE} and DBP15K \cite{JAPE} were extracted from the multilingual DBpedia.
Each setting is an entity alignment task between two cross-lingual versions of DBpedia~\cite{DBpedia}.
Hence, its source and target KGs are from similar sources, resulting in more isomorphic structures.
DWY100K was extracted from DBpedia~\cite{DBpedia}, Wikidata \cite{Wikidata} and YAGO3 \cite{YAGO}.
As YAGO3 also borrows triples from DBpedia, it can be considered as a similar source to DBpedia.
Wikidata is an open KG that both humans and machines can read and edit.
It appears to have fewer isomorphic structures with DBpedia than YAGO3 has.
However, DWY100K, as well as WK3L and DBP15K, lean to structure-based entity alignment, and lack attribute triples to support attribute-enhanced methods.
OpenEA is a recent comprehensive dataset extracted from multilingual DBpedia~\cite{DBpedia}, Wikidata \cite{Wikidata} and YAGO3 \cite{YAGO}.
Its settings from DBpedia and YAGO also have the issue of isomorphic structures.
Furthermore, all of its settings have less heterogeneous attribute values.

%% file: sec3_dataset.tex
\section{\datasetname: Event-centric Entity Alignment Dataset}
This section describes the construction and evaluation of the proposed event-centric entity alignment dataset \datasetname.


\subsection{Dataset Construction}

\subsubsection{Data sources.}
We choose Wikidata \cite{Wikidata} and DBpedia \cite{DBpedia}  as the main sources to build our dataset.
They are two famous and widely used KGs that also contain event entities.
Moreover, the relational structures and attributes of event entities in them are quite heterogeneous, posing challenges to embedding-based alignment methods.
As it is easy to match literals within the same language when comparing the attribute information of entities, we also consider the cross-lingual setting.
We use Wikidata (English) as the source KG and DBpedia (English), DBpedia (French) and DBpedia (Polish) are the target KGs .

\subsubsection{Event entity acquisition.}
Our task requires us to create the source and target KGs based on event entities.
EventKG \cite{EventKG} is primarily responsible for the acquisition of event entities in Wikidata and DBpedia, which contains more than $690,000$ contemporary and historical events.
We identify the event entities in the English versions of Wikidata and DBpedia using the event-entity mappings in EventKG.
Then, we get $50,235$ events in the French version and $31,723$ events in the Polish version with the inter-language links of DBpedia.
These event entity links, along with other entity links, serve as the reference alignment.


\subsubsection{Difficult entity alignment selection.}
As we have mentioned, the literal names of entities may become an opportunistic feature for entity alignment. 
We remove the easy event entity alignment with names that are very similar to avoid the name bias \cite{AttrGNN} and make our dataset challenging enough.
For Wikidata and the English DBpedia, we directly use the string matching method to calculate the similarities of entity names and eliminate the easy alignment pairs whose literal similarities are greater than 0.9.
Finally, we get $20,000$ event alignment pairs.
For the cross-lingual versions (e.g., Wikidata (English) and DBpedia (French)), the literal similarity is relatively low, although the alignment is not difficult.
Hence, we translate their names into English with Google Translate and also select $20,000$ difficult entity pairs whose name similarities are lower than 0.9.

\subsubsection{Entity complement.}
Considering that the spatial information is important for events, we supplement location entities to enrich \datasetname. 
EventKG also stores the event-related location information, mainly through the relations \textit{hasPlace} and \textit{containedInPlace} to associate locations and events.
So, we extract the locations that have the above relations with the acquired events,
and add the corresponding entities for these locations from Wikidata and DBpedia into our dataset.

\subsubsection{Triple complement.}
After obtaining the entities, we can extract triples from their respective data sources for our dataset.
We extract the attribute triples whose head entities are in the obtained entity set. 
The relation triples whose head and tail entities are both in the entity set are also extracted.
Considering that there are richer relation triples between locations and events in EventKG than in Wikidata and DBpedia, 
we also extract these relation triples from EventKG and add some of them to our dataset for triple complement.
Specifically, since these relation triples all come from the same source EventKG, we shuffle them and only add a randomly sampled half of them to each KG to achieve the heterogeneity.

\begin{table}[t!]
	\centering    
	\caption{Statistics of the proposed \datasetname dataset.}
    \label{tab:dataset}
	\setlength{\tabcolsep}{4pt}
	\resizebox{.999\columnwidth}{!}{
	\begin{tabular}{llccccccc}
		\toprule
		\multicolumn{2}{c}{Datasets} & \# Entities & \# Relations & \# Relation trip. & \# Attributes & \# Attribute trip. \\
		\midrule
		\multirow{2}{*}{WD-EN} & Wikidata & $28,877$ & $158$ & $45,193$ & $377$ & $78,680$ \\
		& DBpedia (EN) & $29,842$ & $113$ & $34,953$ & $280$ & $91,795$ \\
		\midrule
		\multirow{2}{*}{WD-FR} & Wikidata & $26,351$ & $68$ & $48,008$ & $442$ & $78,836$ \\
		& DBpedia (FR) & $25,221$ & $101$ & $62,052$ & $281$ & $83,057$ \\
		\midrule
		\multirow{2}{*}{WD-PL} & Wikidata & $25,402$ & $69$ & $42,314$ & $456$ & $82,472$ \\
		& DBpedia (PL) & $25,402$ & $30$ & $33,766$ & $57$ & $56,733$ \\
		\bottomrule
	\end{tabular}}
	\vspace{-5pt}
\end{table}

			    

\subsection{Dataset Evaluation}
We present the statistics of our dataset in Table~\ref{tab:dataset}. 
Next, we analyze the dataset in terms of structural isomorphism and attribute value heterogeneity.

\begin{figure}[!th]
	\centering
	\includegraphics[width=0.8\linewidth]{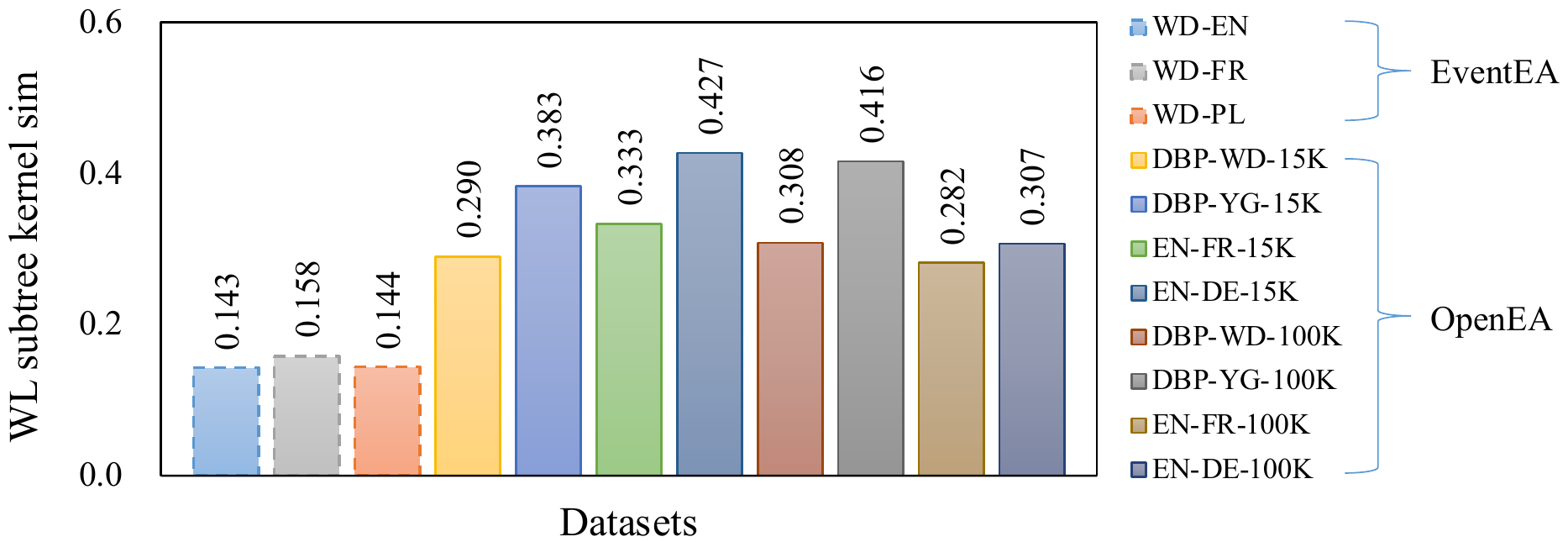}
	\caption{WL subtree kernel similarities of the \datasetname and OpenEA datasets.}
	\label{fig:data_sim}
	\vspace{-10pt}
\end{figure}

\subsubsection{Structural isomorphism.}
The evaluation on the previous entity alignment dataset \cite{openea} only focuses on the properties for a single KG, such as entity degree and clustering coefficient.
However, the alignment task is performed on two KGs, so we mainly consider the structural isomorphism between them, and use this as an indicator to measure the structural similarity of different datasets.
We first simplify two KGs into label graphs, respectively, and delete the isolated entities which cannot utilize the structural information but affect the judgment of structural similarity.
Then, we set the labels of the aligned entities to be the same and we adopt the Weisfeiler-Lehman Subtree Kernels \cite{WL} on the graph to compute the structural similarity between the two KGs.
Fig.~\ref{fig:data_sim} shows the results on \datasetname and OpenEA datasets \cite{openea}.
We can find that the structural similarities of our datasets are low, indicating that they have stronger heterogeneity and it would bring challenges for existing structure-based methods.


\subsubsection{Name heterogeneity.}
\begin{figure}[!t]
	\centering
	\includegraphics[width=0.8\linewidth]{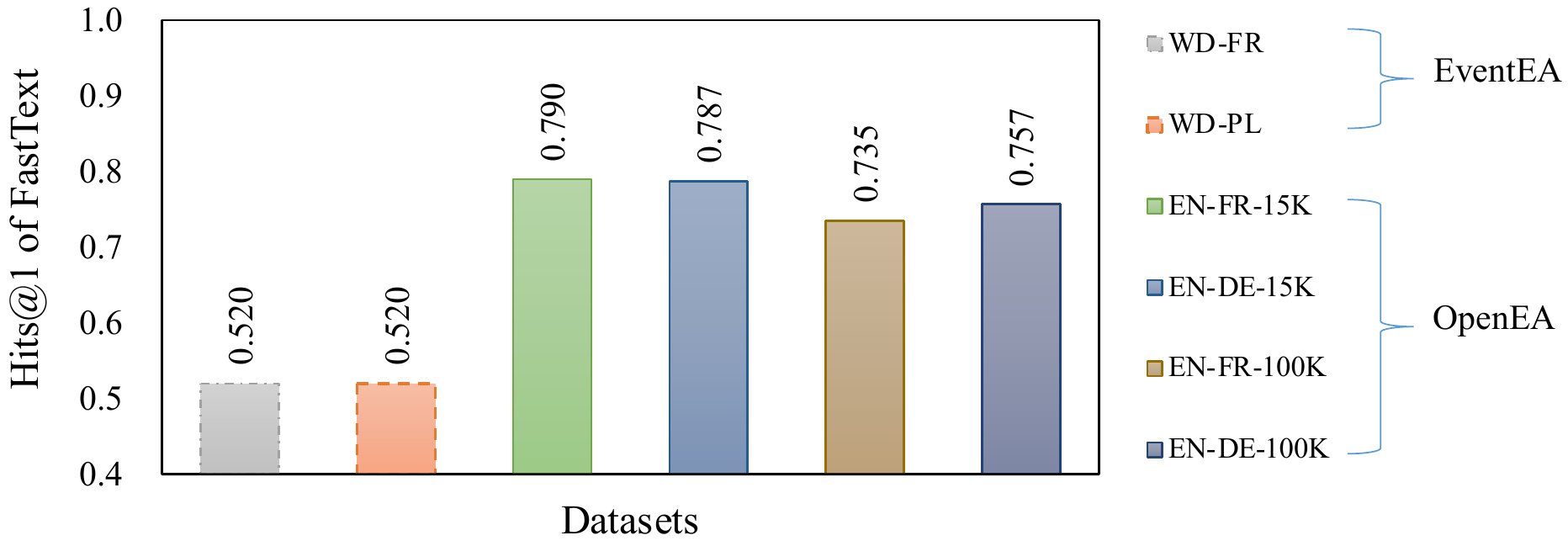}
	\caption{Hits@1 results of FastText on the \datasetname and OpenEA datasets.}
	\label{fig:data_sim_fasttext}
	\vspace{-5pt}
\end{figure}

As we have mentioned, entity names have become the most ``useful'' attribute-related information for entity alignment, which, however, would cause a biased evaluation.
By contrast, in our dataset, entity names have strong heterogeneity.
Hence, it is difficult to align entities through name matching methods.
To further increase the heterogeneity of names and attribute values, 
our \datasetname also contains cross-lingual entity alignment settings,
which makes it difficult to match literals since there is still a large heterogeneity between our attribute values after translation.
To evaluate such heterogeneity, we propose a word embedding based method to see its performance on aligning entity names.
We average the word embeddings of a literal name (translated into English from cross-lingual settings) through FastText \cite{fasttext} as the representations of the entities.
These name-based entity embeddings are used to retrieve entity alignment by nearest neighbor search.
We use Cosine as the embedding similarity measure and calculate the Hits@1 scores.
Fig.~\ref{fig:data_sim_fasttext} shows the results on \datasetname and OpenEA.
We can find that the Hits@1 on \datasetname is much lower than that on OpenEA, illustrating the stronger heterogeneity in entity names of our dataset.



%% file: sec4_method.tex
\section{Time-aware Literal Encoder}
\begin{figure}[!t]
	\centering
	\includegraphics[width=0.95\linewidth]{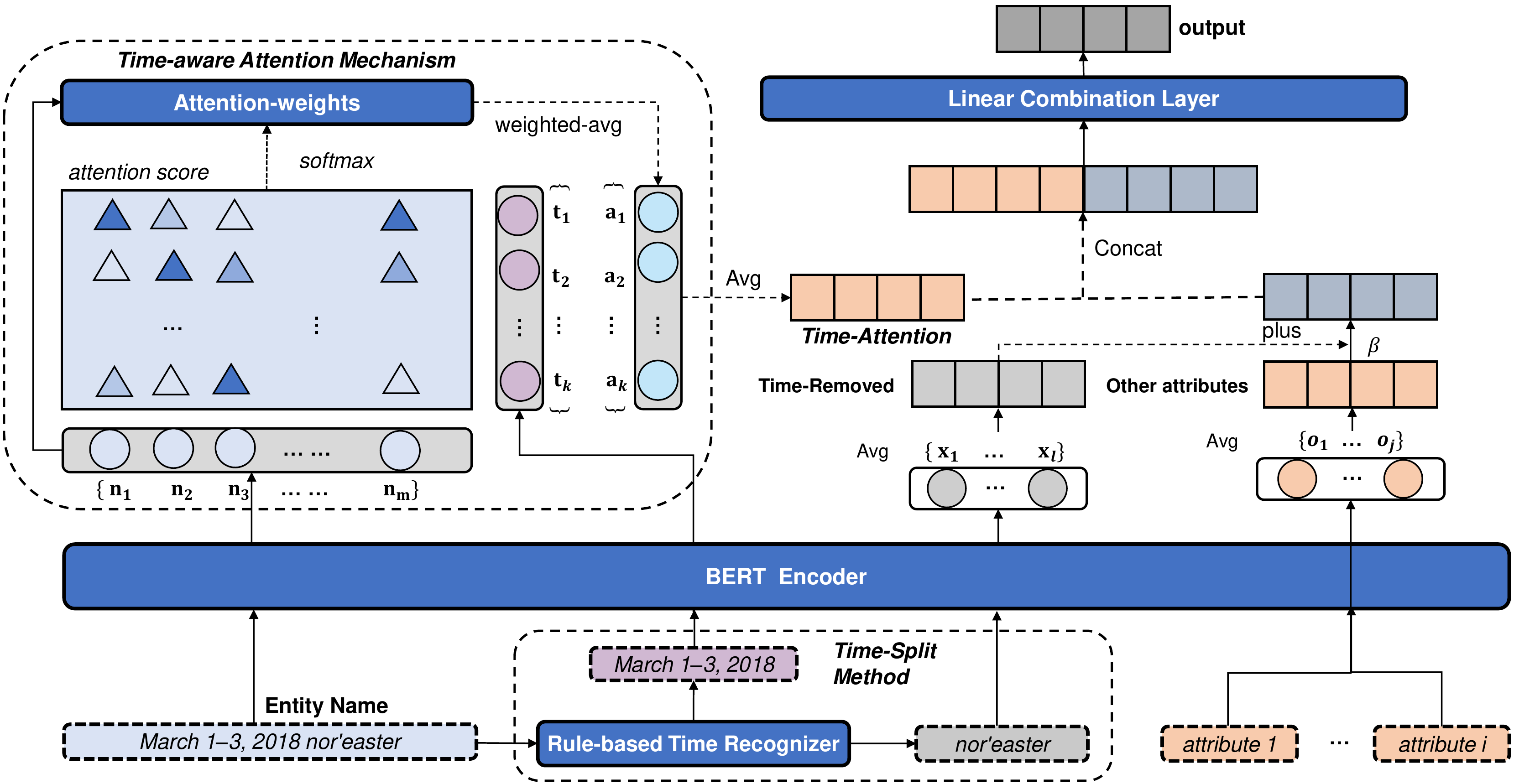}
	\caption{Time-aware literal encoder.}
	\label{fig:encoder}
\end{figure}

Events have strong semantic heterogeneity, i.e., similar names but different semantics.
Moreover, the time information in event names is vital for the alignment.
Therefore, we design a time-aware literal encoder, named TAE, to embed literals while capturing the time information.
The framework is shown in Fig.~\ref{fig:encoder}.

\subsection{Time-split Method}
Given an entity name as input, we use a simple rule-based time recognizer to distinguish the time and non-time information and get the corresponding two string outputs $time$ and $tr$.
We mainly use the regular expression to obtain the time that can be correctly identified in the form of $xxxx$-$xx$-$xx$, $xxxx$-$xx$, $month\ day,\ year$, etc. 
The design of this part can be further refined according to the specific needs of the task, but in our experiments, we found that the simple recognition mechanism has brought good results.

\subsection{BERT Encoder}
We adopt BERT \cite{BERT} as the encoder to obtain the word vectors in the sentence, which can resolve the semantic complexity of event name to some extent.
Specifically, given a $name$, we obtain $\texttt{BERT}(name)=\{{\mathbf{n}_i}\}_{i=1}^m$, where $\mathbf{n}_i$ represents the vector representation of the $i$th word and $m$ is the length of the sequence. 
Similarly, for $time$ and $tr$, we have $\texttt{BERT}(time)=\{{\mathbf{t}_i}\}_{i=1}^k$ and $\texttt{BERT}(tr)=\{{\mathbf{ x}_i}\}_{i=1}^l$.
For other attribute values 
of an entity, we concatenate the texts together with spaces to obtain a long attribute value string $attr$ and obtain $\texttt{BERT}(attr)=\{\mathbf{o}_i\}_{i=1}^j$.


\subsection{Time-aware Attention Mechanism}
We use the attention mechanism to obtain time-aware attention vector representations.
Specifically, $\{\mathbf{t}_i\}_{i=1}^k$ is used as the query sequence, 
and  $\{\mathbf{n}_i\}_{i=1}^m$ is used as the key and value sequence.
We first compute an attention distribution over $\{\mathbf{n}_i\}_{i=1}^m$ for each $\mathbf{t}_{i}$ which can be formalized as follows:
\begin{equation}
\alpha_{ij}={\rm softmax}({\cos}(\mathbf{n}_j,\mathbf{t}_i))=\frac{{\exp}\big({\cos}(\mathbf{n}_j,\mathbf{t}_i)\big)}{\sum_{k=1}^m {\exp}\big({\cos}(\mathbf{n}_k,\mathbf{t}_i)\big)},
\end{equation}
where $\rm{cos(\cdot,\cdot)}$ denotes the cosine similarity function.
Then, we use the weighted sum of $\{{\mathbf{n}_i}\}_{i=1}^m$ through the attention scores to obtain the attention representation for each $\mathbf{ t}_i$ over the entire name, which is defined as follows: 
\begin{equation}
\mathbf{a}_i=\sum_{j=1}^m\alpha_{ij}\mathbf{n}_j.
\end{equation}




\subsection{Linear Combination Layer}
To enhance the effect of time information, we average the set of time attention representations $\{\mathbf{ a }_i\}_{i=1}^k$ and get the final time attention embedding $\mathbf {h}$ as follows:
\begin{equation}
\mathbf{h}=\frac{1}{k}\sum_{i=1}^k\mathbf{a}_i.
\end{equation}
We also average the word vectors $\{\mathbf{ x}_i\}_{i=1}^l$ and attribute value vectors $\{{\mathbf{ o}_i}\}_{i=1}^j$ to obtain the time-removed name embedding $\mathbf{ {r}}$ and the aggregated embedding $\mathbf{ {g}}$ of other attributes, respectively.
Then, we make a simple fusion between 
them:
\begin{equation}
\mathbf{f}=\mathbf{r} + \beta * \mathbf{g},
\end{equation}
where $\beta$ is a hyper-parameter for balance. 
Finally, we concatenate $\mathbf{{h}}$ and $\mathbf{{f}}$ and apply a linear combination layer to obtain the complete time-aware encoding:
\begin{equation}
\rm{TAE(attribute\ values)=\mathbf{W_{comb}}[\mathbf {h;f}]+\mathbf{b_{comb}}},
\end{equation}
where $\mathbf{W_{comb}} \in {\mathbb{R}^{d\times 2d}}$ is a weight matrix, $\mathbf{ b_{comb}} \in {\mathbb{R}^d}$ is a bias vector.




\subsection{Alignment Learning Loss}
We choose the contrastive alignment loss which is widely used by existing work:
\begin{equation}
\mathcal{L} = \sum_{(i,j) \in P}||\,\mathbf{e}_i - \mathbf{e}_j\,|| + \sum_{(i',j') \in N}\Big[\lambda - ||\,\mathbf{e}_{i'} - \mathbf{e}_{j'}\,||\Big]_+,
\end{equation}
where $P$ is the set of pre-aligned entity pairs and $N$ is the set of non-aligned entity pairs obtained by randomly replacing one entity in the pre-aligned two entities.
$||\cdot||$ denotes the L2-norm of the vector, $[\cdot]_+=\max(0,\cdot)$, and $\lambda$ is a hyperparameter.
The loss function expects the distance between the representations of two aligned entities to be small while non-aligned ones to be large (i.e., greater than $\lambda$).

%% file: sec5_experiment.tex
\section{Evaluation}
In this section, we report the experimental results on our dataset.

\subsection{Experiment Settings}

\subsubsection{Environment.}
We run experiments on a workstation with two Intel Xeon Gold 5122 3.6GHz CPUs, a NIVIDA GeForce RTX 3090 GPU and Ubuntu 18.04 LTS.

\vspace{-5pt}
\subsubsection{Evaluation protocol.}
In our experiments, 
the default alignment direction is from left to right,
i.e.,
we use Wikidata as the source and align it to DBpedia. 
To retrieve entity alignment, the cosine similarity is used to measure the similarity between two entity embeddings.
Following the convention, 
we employ Hits@$k$ ($k = 1, 10$) and mean reciprocal rank (MRR) as the evaluation metrics.

\subsubsection{Baseline methods.}
We divide the baseline methods into five groups:

\begin{itemize}

\item \textbf{Using the string similarity to match entity names.} 
These methods are based on the string similarity of entity names.
We consider the Levenshtein related methods taken from the \texttt{python-Levenshtein} 0.12.2 package,\footnote{\url{https://pypi.org/project/python-Levenshtein/}} 
and the difflib.SequenceMatcher method in the \texttt{difflib} package from Python.

\smallskip \item \textbf{Encoding entity names into vectors to match.}
These methods use word embeddings or pre-trained language models to encode entity names as entity representations for alignment search.
We consider FastText \cite{fasttext} and BERT (bert-base-uncased) \cite{BERT}.
As for BERT, it has three variants: BERT (embed), BERT (L4-avg), and BERT (L4-concat), 
denoting using the output of the embedding layer, averaging the representations of the last four layers, and concatenating the representations of the last four layers, respectively.

\smallskip \item \textbf{Structure-based entity alignment methods.}
These methods learn from relation triples for entity alignment.
We consider 14 structure-based entity alignment methods (including some variants of conventional KG embedding models) in OpenEA \cite{openea}, as listed in the third group of Table~\ref{tab:ent_alignment}.


\smallskip \item \textbf{Attribute-enhanced methods.}
These methods incorporate entity attributes for embedding learning.
We consider 5 attribute-enhanced methods in the fourth group of Table~\ref{tab:ent_alignment}.
We use the implementations in OpenEA~\cite{openea}.

\smallskip \item \textbf{Our method.}
TAE uses the word vectors obtained by BERT with our proposed time-aware encoder to embed entity names for alignment search.
\end{itemize}

It is worth noting that, in the cross-lingual settings, we preprocess non-English entity names through Google Translate.
The hyper-parameter settings of our method are reported in the appendix of supplemental materials, 
while the settings of baseline methods can be seen in the source code.

\subsection{Main Results and Analyses}
Table~\ref{tab:ent_alignment} presents the results of Hits@1, Hits@10 and MRR for different methods on \datasetname.
In general,
our method TAE achieves the best results,
and the structure-based methods fail to achieve promising performance.
We conduct in-depth and comprehensive analyses 
of the results from three aspects:

\subsubsection{String similarity vs. embedding similarity.}
It can be observed that
FastText achieves the best results on all metrics compared to the seven baselines that only 
using entity names for alignment.
Also, the results of encoding using the BERT embedding layer can be ranked in the top three on all datasets.
It is enough to show that 
the embedding technique is more powerful than the traditional string matching methods.
In addition, we find that the performance of string matching in the cross-lingual settings is much worse than that in the monolingual settings.
This indicates that the cross-lingual settings in our dataset have strong heterogeneity of attribute values even after machine translation.
We also notice that such heterogeneity has less effect on the methods using word vectors, showing the advantage of word vectors in capturing the latent semantics of names.
Meanwhile, we also notice that the performance of FastText is better than some common BERT-based methods, 
which shows that pre-trained language models is not always a better choice for encoding entity names.
This is probably related to the reason that no fine-tuning is performed.
These results demonstrate the effectiveness of embedding techniques in comparing literal names.


\begin{table}[!t]
	\centering
	\caption{Entity alignment results on our datasets.}
	\label{tab:ent_alignment}
	\resizebox{1.0\textwidth}{!}{
		\begin{threeparttable}
		\setlength{\tabcolsep}{0.3em}
		\begin{tabular}{lccccccccc}
		\toprule
		& \multicolumn{3}{c}{WD-EN} & \multicolumn{3}{c}{WD-FR} & \multicolumn{3}{c}{WD-PL} \\
		\cmidrule(lr){2-4} \cmidrule(lr){5-7} \cmidrule(lr){8-10} 
		& Hits@1 & Hits@10 & MRR & Hits@1 & Hits@10 & MRR & Hits@1 & Hits@10 & MRR \\ 
		\midrule
		Levenshtein-Ratio & .580 & .801 & .661 & .394 & .499 & .431 & .361 & .534 & .419 \\		
		Levenshtein-Jaro  & .619 & .817 & .693 & .391 & .483 & .425 & .341 & .497 & .394 \\
		Levenshtein-Jaro-Winkler    & .556 & .755 & .630 & .339 & .483 & .388 & .267 & .378 & .308 \\
		SequenceMatcher & .476 & .705 & .557 & .418 & .537 & .461 & .474 & .660 & .539 \\
		\midrule
		FastText~\cite{fasttext}        & .657 & .866 & .734 & .536 & .719 & .598 & .541 & .723 & .606 \\
		BERT (embed)~\cite{BERT}        & .581 & .820 & .668 & .508 & .665 & .562 & .524 & .700 & .583 \\
		BERT (L4-avg)~\cite{BERT}       & .505 & .721 & .579 & .523 & .694 & .582 & .467 & .663 & .536 \\
		BERT (L4-concat)~\cite{BERT}    & .499 & .713 & .573 & .536 & .719 & .598 & .464 & .659 & .532 \\ 
	    \midrule
		$\text{MTransE}^\triangle$~\cite{MTransE} & .053 & .128 & .080 & .106 & .223 & .150 & .053 & .110 & .061 \\
		$\text{TransH}^\triangle$~\cite{TransH} & .045 & .097 & .064 & .080 & .137 & .103 & .036 & .072 & .050 \\
		$\text{TransD}^\triangle$~\cite{TransD} & .042 & .086 & .058 & .080 & .138 & .104 & .044 & .081 & .059 \\
		$\text{HolE}^\triangle$~\cite{HolE} & .004 & .017 & .074 & .015 & .037 & .002 & .000 & .003 & .002 \\
		$\text{ProjE}^\triangle$~\cite{ProjE} & .023 & .061 & .037 & .063 & .112 & .083 & .007 & .017 & .011 \\
		$\text{IPTransE}^\triangle$~\cite{IPTransE} & .047 & .100 & .066 & .084 & .141 & .106 & .043 & .085 & .060 \\
		$\text{ConvE}^\triangle$~\cite{ConvE} & .016 & .048 & .027 & .048 & .087 & .064 & .003 & .011 & .007 \\
		$\text{AlignE}^\triangle$~\cite{BootEA} & .177 & .410 & .225 & .142 & .293 & .195 & .092 & .190 & .128 \\
		$\text{BootEA}^\triangle$~\cite{BootEA} & .219 & .485 & .309 & .168 & .325 & .225 & .104 & .251 & .155 \\
		$\text{SimplE}^\triangle$~\cite{SimplE} & .002 & .009 & .005 & .008 & .029 & .016 & .002 & .010 & .006 \\
        $\text{RotatE}^\triangle$~\cite{RotatE} & .047 & .137 & .077 & .075 & .196 & .117 & .035 & .101 & .058 \\
		$\text{SEA}^\triangle$~\cite{SEA} & .156 & .389 & .235 & .120 & .273 & .177 & .087 & .200 & .127 \\
		$\text{RSN4EA}^\triangle$ ~\cite{RSN4EA} & .203 & .350 & .254 & .164 & .239 & .191 & .107 & .161 & .127 \\
		$\text{AliNet}^\triangle$~\cite{AliNet} & .146 & .287 & .196 & .150 & .234 & .184 & .098 & .167 & .126 \\
		\midrule
		$\text{JAPE}^\triangle$~\cite{JAPE} & .048 & .126 & .074 & .102 & .199 & .139 & .049 & .092 & .065 \\ 
		$\text{GCNAlign}^\triangle$~\cite{GCN-Align} & .058 & .144 & .088 & .099 & .245 & .154 & .058 & .139 & .087 \\
		$\text{AttrE}^\triangle$~\cite{AttrE} & .165 & .337 & .224 & .135 & .267 & .185 & .074 & .164 & .107 \\
		$\text{IMUSE}^\triangle$~\cite{IMUSE} & .395 & .484 & .426 & .194 & .286 & .231 & .174 & .249 & .202 \\
		$\text{MultiKE}^\triangle$~\cite{MuiltiKE} & .230 & .404 & .289 & .144 & .253 & .181 & .132 & .206 & .158 \\
		\midrule
		TAE (ours) & \textbf{.769} & \textbf{.917} & \textbf{.825} & \textbf{.826} & \textbf{.905} & \textbf{.857} & \textbf{.792} & \textbf{.896} & \textbf{.831} \\
		\bottomrule
		\end{tabular}
		$\triangle$ Results are produced by ourselves using OpenEA~\cite{openea}. Best results are marked in boldface.
	    \end{threeparttable}}
\end{table}

\subsubsection{Structure-based vs. attribute-enhanced.}
Generally, the structure-based methods that only use relation triples perform comparatively poorly. 
This is because our \datasetname has fewer isomorphic structures that structure-based methods heavily rely on.
Among these methods, more advanced KG embedding methods do not bring significant improvement, e.g., TransH, TransD, HolE, ProjE, ConvE, SimplE and RotatE. 
This is because that \datasetname has relatively few relation triples which may raise the overfitting issue for the KG embedding methods.
Meanwhile, it is worth noting that semi-supervised strategies are also beneficial on \datasetname according to the results of SEA and BootEA. 
In addition, extending the relation triples to paths or utilizing neighborhood subgraphs also bring some improvement, e.g., RSN4EA and AliNet.
For the methods that use additional attribute information, JAPE and GCN-Align do not perform well because they only use attribute correlations but neglect literals.
This indicates the importance of introducing attribute values information for \datasetname.
IMUSE performs the best among the attribute-enhanced methods.
This is because IMUSE deals with the numerical values during the data preprocessing phase which can utilize many time-related attributes in \datasetname.
MultiKE also performs well owing to its advantage of combining multiple views of entities.
We also notice that AttrE, which also uses attribute values, obtains an underwhelming performance.
It is because that AttrE needs to align the predicates, while the schema heterogeneity in \datasetname is huge.
In summary, both structure-based and attribute-enhanced methods fail to achieve their claimed performance on \datasetname, which calls for more robust solutions to embedding-based entity alignment.

\subsubsection{Analyses of TAE.}
Compared with the string matching methods, TAE uses a pre-trained language model to encode literals, which can help resolve the literal heterogeneity and capture the latent semantic associations between two aligned entity names.
Moreover, different from directly using pre-trained word embeddings, we design a time-aware mechanism so that the entities with complex semantic names can also have a high similarity with their aligned entities.
TAE also incorporates other attribute information, providing additional information for entities with ambiguous names.
However, TAE does not use the structural information, which makes it limited and unable to align entities with few attributes, ambiguous or overly complex semantic names.

\begin{table}[!t]
	\centering
	\caption{Ablation study of TAE.}
	\label{tab:ablation}
	\resizebox{1.0\textwidth}{!}{
		\begin{threeparttable}
			\setlength{\tabcolsep}{5pt}
			\begin{tabular}{lccccccccc}
				\toprule
				& \multicolumn{3}{c}{WD-EN} & \multicolumn{3}{c}{WD-FR} & \multicolumn{3}{c}{WD-PL} \\
				\cmidrule(lr){2-4} \cmidrule(lr){5-7} \cmidrule(lr){8-10} 
				& Hits@1 & Hits@10 & MRR & Hits@1 & Hits@10 & MRR & Hits@1 & Hits@10 & MRR \\ 
				\midrule
			    TAE & 
			    \textbf{.769} & \textbf{.917} & \textbf{.825} & \textbf{.826} & \textbf{.905} & \textbf{.857} & \textbf{.792} & \textbf{.896} & \textbf{.831} \\
			    TAE w/o TA &
			    .595 & .834 & .679 & .413 & .678 & .497 & .447 & .743 & .544 \\
			    TAE w/o OA &
			    .767 & \textbf{.917} & \textbf{.825}& .820 & .904 & .852 & .788 & .894 & .828 \\
				\bottomrule
			\end{tabular}
	\end{threeparttable}}
\end{table}

\subsection{Ablation Study}
Table~\ref{tab:ablation} shows the results of ablation studies on TAE.
TA and OA denote the time-aware mechanism and other attribute values except names, respectively.
The time-aware mechanism contributes a lot to the improvement, showing that the consistency in time is of great help in solving the heterogeneity of event names.
It also reflects that, in addition to the time information, the heterogeneity of attribute values is very large, making our method ineffective.
We find that fusing other attributes only brings a little improvement, indicating that simply processing strongly heterogeneous attribute values cannot achieve good results on \datasetname,
which calls for more efforts to resolve the attribute heterogeneity.

\subsection{More Analyses}
\subsubsection{Recall of aligning different types of entities.}
\begin{figure}[!t]
	\centering
	\includegraphics[width=\linewidth]{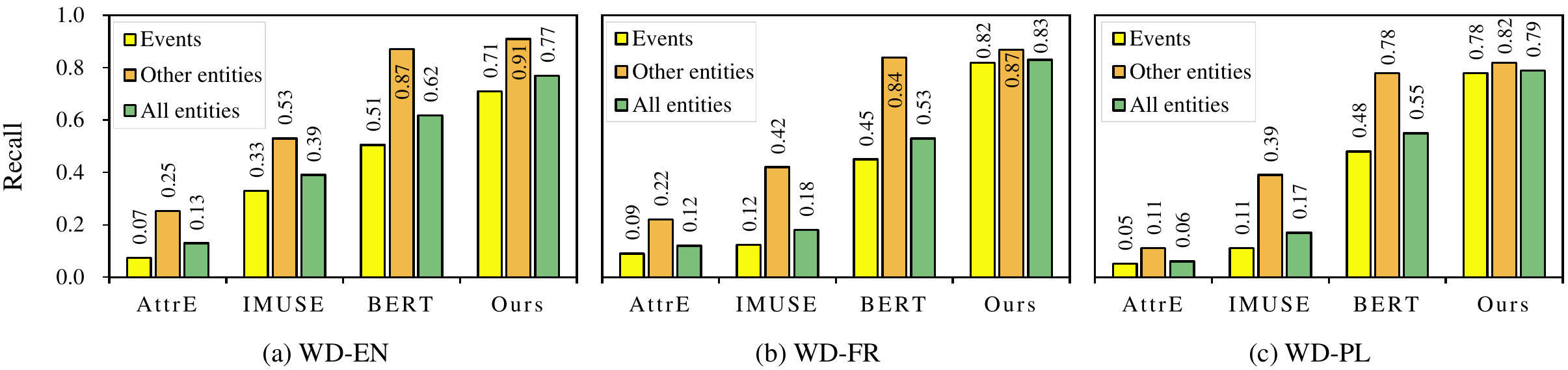}
	\caption{Recall of aligning events, other entities and all entities.}
	\label{fig:more_analyse}
\end{figure}

Fig.~\ref{fig:more_analyse} shows the recall of aligning events, other entities, and all entities on our datasets, respectively.
We can find that all methods obtain the lowest recall on events across all datasets, which indicates that aligning such special entities with strong heterogeneity is challenging.
The results further confirm that the aforementioned biases exist, as the alignment results on other entities are obviously better.
We also notice that the obtained recall of TAE on events and other entities is nearly close, which demonstrates that our proposed method is suitable for such a setting with weak structural isomorphism and strong attribute heterogeneity.

\begin{table}[!t]
	\centering
	\caption{Cases of entity alignment that our method correctly predicts.
	The second column lists the top-3 nearest DBpedia neighbors of each Wikidata entity in the first column, while the last three columns show the similarities obtained by different methods. }
	\label{tab:correct_cases}
	\resizebox{1.0\textwidth}{!}{
	\begin{threeparttable}
	\setlength{\tabcolsep}{5pt}\renewcommand\arraystretch{1.2}
    \begin{tabular}{llccc} 
    \toprule
    Source entities & Top-3 target entities & TAE & Leven. & BERT \\  
    \midrule
    \multirow{3}{3cm}{1948–49 Svenska m\"{a}sterskapet} & 1948–49 Svenska m\"{a}sterskapet (men's handball) & .874 & .767 & .825 \\
    & 1902 Svenska M\"{a}sterskapet & .822 & .830 & .915 \\
    & 1949–50 Svenska m\"{a}sterskapet (men's handball) & .812 & .685 & .795 \\
    \midrule
    & 1997 Finnish Football Championship & .922 & .385 & .533 \\
    1997 Veikkausliiga & 1997 Norwegian Football Championship & .822 & .407 & .516 \\
    & 1996 Finnish Football Championship & .813 & .346 & .513 \\
    \midrule
    & 2019 Southern Yemen clashes & .754 & .261 & .697 \\
    Battle of Aden 2019 & Battle of Fujian & .750 & .629 & .789 \\
    & Battle of M\d{a}o Kh\^{e} & .748 & .611 & .687 \\
    \bottomrule
    \end{tabular}
    \end{threeparttable}
}
\end{table}

\begin{table}[!t]
	\centering
	\caption{Wrong cases.
	The second and fourth columns list the wrongly predicted nearest DBpedia entity and the ground-truth counterpart for each source entity in the first column, respectively.
    Their similarities obtained 
    through our method are listed in the third and fifth columns, respectively.}
	\label{tab:wrong_cases}
	\resizebox{1.0\textwidth}{!}{
	\begin{threeparttable}
	\setlength{\tabcolsep}{0.3em}
	\renewcommand\arraystretch{1.2}
    \begin{tabular}{p{4.5cm}p{4cm}clc} 
    \toprule   
    Source entities & Top-1 target entities & TAE & Gold target entities & TAE \\  
    \midrule
    Secure Electronic Network for Travelers Rapid Inspection& Armed Forces Association Cycling Classic & .632 
    & SENTRI & .261 \\
    \midrule
    {Black May} & Black May (1943) & .878 
    & Black May (1992) & .853 \\
    \midrule
   {1979 Nice events} & Brighton Tennis Tournament (WTA 1979) & .821 
    & Nice airport tsunami & .427 \\
    \bottomrule
    \end{tabular}
    \end{threeparttable}
}
\end{table}

\subsubsection{Case study.}
Table~\ref{tab:correct_cases} shows some correct cases predicted by our method.
The first case demonstrates that our time-aware mechanism can capture the time information accurately in names, 
while the second and third cases show that our method can also capture complex latent semantics.
Moreover, the methods based on string similarities or BERT-encoded similarities fail to capture those two kinds of information.
We also give some wrong cases made by our method in Table~\ref{tab:wrong_cases}.
In the first case, the correct target entity name is so short for the source entity name that it is hard to capture their similarities.
The second case demonstrates the importance of injecting other information to assist the judgment of entity alignment when the source entity name is contained in the names of the candidate target entities.
The third case shows that our method fails to align identical entities when their surface names are quite different. 




%% file: sec6_conclusion.tex
\section{Conclusions and Future Work}
In this paper,
we propose an event-centric dataset, \datasetname, with weak structural isomorphism and strong attribute heterogeneity,
which greatly eliminates the biases of existing datasets for entity alignment. 
Furthermore, we evaluate existing representative embedding-based methods on \datasetname, 
and the lower-than-expected results of these methods suggest that we need more robust methods. 
In addition, for \datasetname, we design a time-aware literal encoder for the heterogeneity of attribute values, which achieves promising results. 

In future work, we will start from the following aspects:
\begin{itemize}

\smallskip \item \textbf{Better structure learning.}
Current structure-based methods are not ideal on \datasetname. 
We should consider how to better learn the associations between strongly heterogeneous graphs based on the existing experimental results.

\smallskip \item \textbf{Better method.}
Although TAE has achieved good results,
according to the analyses of experimental results, there are still some entities whose literal semantics are too complex to solve well. 
We should consider how to combine structural information for better method design.

\smallskip \item \textbf{More general.}
To make our event-centric dataset more universal, 
we should consider increasing the diversity of entities
while maintaining the weak structural isomorphism and strong attribute heterogeneity of the dataset.

\smallskip \item \textbf{Larger size.}
We can increase the size of \datasetname, allowing us to more accurately evaluate the effect of different methods on such challenging dataset.
\end{itemize}